\documentclass[runningheads,a4paper]{llncs}
\usepackage{amsmath}
\usepackage{amssymb}
\usepackage{graphicx}
\usepackage{booktabs}
\usepackage{listings}
\usepackage{xcolor}
\usepackage{hyperref}
\usepackage{caption}
\usepackage{subcaption}

\usepackage{url}
\usepackage{color}
\usepackage{cite}
\usepackage{cleveref}
\usepackage{multirow}
\newcommand{\orcid}[1]{\href{https://orcid.org/#1}{\includegraphics[scale=0.02]{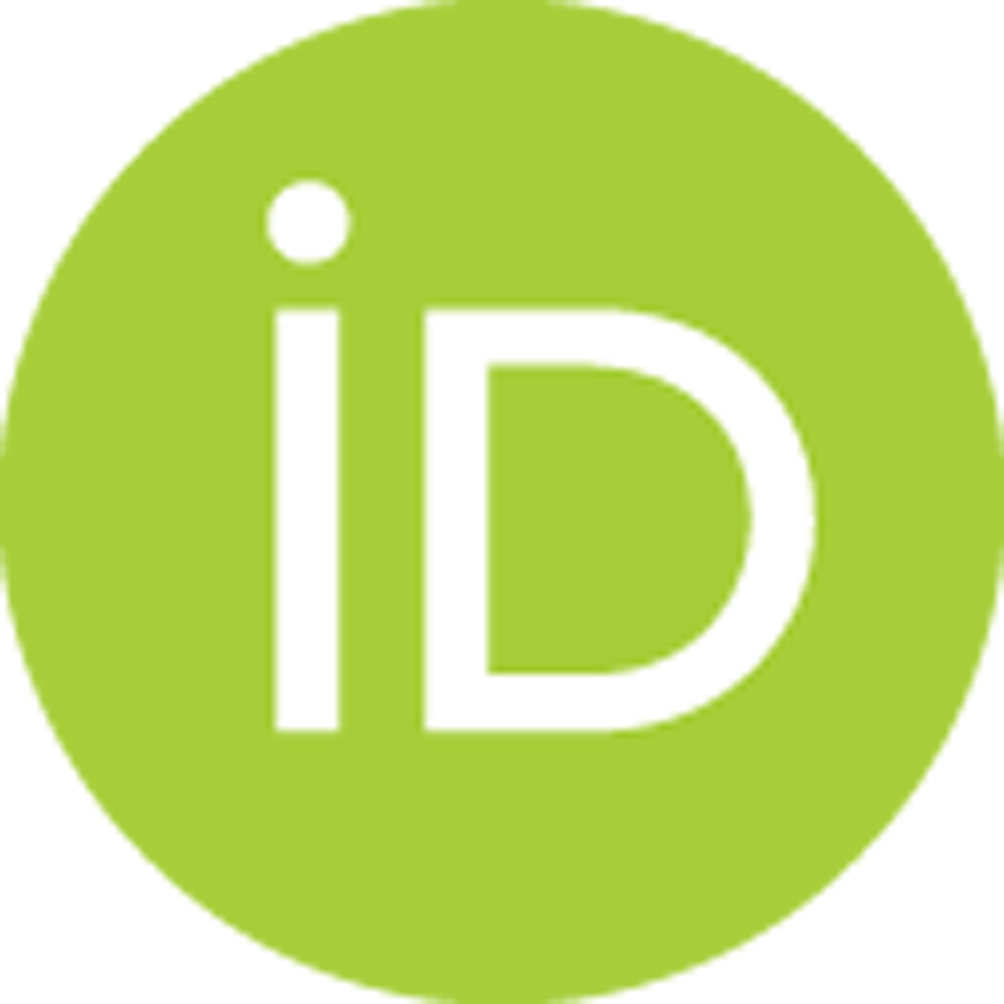}}}

\hypersetup{
    colorlinks=true
}
\title{The Federated Tumor Segmentation and Classification (FeTS++) Challenge}
\titlerunning{FeTS++ Challenge}
\author{
Wenyi Tang\inst{1,2,3,\orcid{0000-0002-2655-8884}}
\and
Carla Pitarch-Abaigar\inst{1,2,3,\orcid{0000-0002-6015-244X}}
\and
Siddhesh Thakur\inst{1,2,3,\orcid{0000-0003-4807-2495}}
\and
Verena Chung\inst{4,\orcid{0000-0002-5622-7998}}
\and
Konstantinos Koukoutegos\inst{1,2,3,\orcid{0000-0003-4392-4476}}
\and
Alexandros Karargyris\inst{5,\orcid{0000-0002-1930-3410}}
\and
Holger R. Roth\inst{6,\orcid{0000-0002-3662-8743}}
\and
Spyridon Bakas\inst{1,2,3,5,7,8,*,\orcid{0000-0001-8734-6482}}
}

\authorrunning{Tang et al.}

\institute{\scriptsize{
Research Center for Federated Learning in Medicine, Indiana University School of Medicine, Indianapolis, IN, USA.
\and
Division of Computational Pathology, Department of Pathology and Laboratory Medicine, Indiana University School of Medicine, Indianapolis, IN, USA.
\and
Indiana University Melvin and Bren Simon Comprehensive Cancer Center, Indianapolis, IN, USA.
\and
Sage Bionetworks, Seattle, WA, USA.
\and
MLCommons, Dover, DE, USA.
\and
NVIDIA, Santa Clara, CA, USA.
\and
Departments of Radiology and Imaging Sciences; Neurological Surgery; Biostatistics and Health Data Science, Indiana University School of Medicine, Indianapolis, IN, USA.
\and
Department of Computer Science, Luddy School of Informatics, Computing, and Engineering, Indiana University, Indianapolis, IN, USA}
\\
\textsuperscript{*} Corresponding author: \email{\{spbakas@iu.edu\}}}

\begin{document}
\mainmatter
\maketitle
\setcounter{footnote}{0} 
\begin{abstract}

International challenges have become the standard for developing and benchmarking biomedical image analysis methods using centralized multi-institutional data.
Federated learning (FL) describes a paradigm-shift on multi-institutional collaborations, where models are trained across institutions while their data are always retained locally. The Federated Tumor Segmentation (FeTS) challenge, first held in 2021, was the first challenge dedicated to FL in medical imaging and editions in 2022 and 2024 pushed the frontiers of real-world method validation.
FeTS++ 2027 returns after a two-year gap with a substantially larger and more diverse cohort, and with the two "pluses" that name the challenge: an official classification task alongside segmentation, and a move of the infrastructure to NVIDIA FLARE (NVFlare).
Built on the clinically acquired, multi-institutional multi-parametric MRI data of the BraTS 2025 Lighthouse challenge, FeTS++ spans gliomas and meningiomas across patient cohorts from institutions in North America, Europe, Asia, and Africa, and focuses on ranking submitted methodological approaches according to their model utility and an efficiency-aware score.
Task 1 asks participants to find the best strategy for aggregating knowledge from per-institution segmentation models, while Task 2 asks them to distinguish glioma from meningioma under the same federated protocol, in which the entity label is heavily confounded with the site of origin.

\end{abstract}


\section{Introduction}
Gliomas and meningiomas are the most common primary brain tumors in adults. Gliomas are the most common malignant tumors of the central nervous system (CNS), whereas meningiomas are the most common intracranial tumors. Glioblastoma, representing the most aggressive form of glioma, accounts for more than half of all primary malignant CNS tumors~\cite{Price2025} and is characterized by extreme intrinsic heterogeneity in appearance, shape, and histology~\cite{baid2021}. Despite standard of care comprising surgical resection followed by radiotherapy and temozolomide, median overall survival remains around 15 months~\cite{Stupp2005}. This poor prognosis is largely due to its diffusely infiltrative growth beyond the radiologically visible margins, which makes tumor recurrence almost inevitable. Meningiomas on the other hand, the most common primary intracranial tumors in adults, represent over 40\% of all primary brain and other CNS tumors~\cite{Price2025}. Approximately 80\% are CNS WHO grade 1 tumors that are typically well controlled with observation, surgical resection, and/or radiotherapy. However, grade 2 and 3 meningiomas are associated with significantly higher morbidity and mortality and often recur despite optimal management~\cite{labella2023}. 

Multiparametric magnetic resonance imaging (mpMRI), typically comprising T1-weighted (T1), post-contrast T1-weighted (T1-Gd), T2-weighted (T2), and T2-FLAIR sequences, is the standard of care for diagnosis, surgical and radiotherapy planning, and response assessment. Accurate delineation of tumor sub-regions, including enhancing tumor, necrotic tumor core, and peritumoral edematous tissue, is essential for patient management, yet manual segmentation remains time-consuming and subject to inter-rater variability. 

The Brain Tumor Segmentation (BraTS) challenge has served since 2012 as the community benchmark for automated brain tumor analysis on mpMRI~\cite{menze2015,bakas2019}. Since 2023, BraTS has evolved into a cluster of challenges covering tumors beyond glioma, including meningioma, brain metastases, and pediatric brain tumors~\cite{labella2023,BraTS2025}, as well as beyond segmentation workloads changing the definition of BraTS to `Brain TumorS'. Nevertheless, BraTS relies on centralized data collection. Data sharing across institutions is hindered by privacy regulations such as Health Insurance Portability and Accountability (HIPAA) and the European Union General Data Protection Regulation (GDPR), along with data ownership and legal considerations, which limit the size and diversity of the datasets available for developing robust and generalizable algorithms~\cite{baid2021}.

Federated Learning (FL) offers a solution to this limitation by enabling multiple institutions to collaboratively train a shared model without exchanging patient data. Only model updates are communicated, and the patient data are retained within their local site~\cite{sheller2020federated,Rieke2020}. Building on this paradigm, the Federated Tumor Segmentation (FeTS) initiative showed that FL can be used in practice to train a glioblastoma segmentation model across dozens of institutions worldwide~\cite{pati2022federated}. Within this initiative, the MICCAI FeTS challenge was launched in 2021 as the very first FL challenge across fields. The MICCAI FeTS challenge focused on benchmarkin FL aggregation and segmentation algorithms on medical imaging data distributed across multiple international institutions~\cite{Zenk2025, linardos2025miccai}.

The Federated Tumor Segmentation and Classification (FeTS++) Challenge 2027 is the latest edition of the FeTS series, and its two pluses mark two extensions: i) a tumor classification workload added alongside segmentation, and ii) a migration of the framework infrastructure to NVIDIA FLARE (NVFlare) \cite{roth2022nvidia}. The challenge therefore spans two official tasks under one federated learning benchmark: federated brain tumor segmentation (Task 1) and federated brain tumor classification (Task 2). Where earlier editions focused on glioblastoma, FeTS++ 2027 covers two tumor entities, namely gliomas and meningiomas, and draws on a substantially larger and more diverse clinically acquired, multi-institutional dataset of over 3,500 patients from the BraTS 2025 Lighthouse Challenge. Participants develop weight aggregation strategies and client training behavior, whereas the model architecture is fixed so that gains are attributable to the FL aggregation method. Submitted methods will be evaluated on hidden data in a controlled NVFlare simulator that respects the FL paradigm, and will be ranked with an efficiency-aware score using a statistically-aware ranking for fair Benchmarking of AI Models (namely PermRanker) \cite{permranker}.

In summary, FeTS++ brings together to the scientific community:
\begin{itemize}
    \item A federated brain-tumor benchmark of over 3,500 patients spanning gliomas, meningiomas, and multiple patient populations, including a Sub-Saharan Africa cohort.
    \item Two official tasks under one FL protocol, with locked model architectures so that gains are attributable to the FL aggregation method.
    \item A reproducible NVFlare-/MONAI-based simulator in which public and hidden evaluation share one code path and one frozen evaluator.
    \item An efficiency-aware ranking protocol that combines accuracy with communication cost and pairwise statistical testing.
\end{itemize}

\section{Materials \& Methods}

\subsection{Data} 
\label{sec:data}

  \subsubsection{Data Sources}
    This challenge builds on the BraTS 2025 Lighthouse data \cite{BraTS2025}, publicly released and augmented with a de-identified institution identifier for each training case. It combines three sources: glioma cases carried over from the FeTS 2024 partition, meningioma cases from the meningioma challenge cohort, and a Sub-Saharan Africa cohort \cite{labella2024,labella2023,adewole2025}. Each case contains clinically acquired multi-parametric structural MRI at a single timepoint, with four co-registered sequences: native T1, T1-Gd, T2, and T2-FLAIR, together with expert reference standard sub-region labels for the training cases. 

    For this FeTS iteration, challenge datasets are described in the Croissant standard format \cite{akhtar2024croissant}. Croissant is an open metadata format for machine learning datasets maintained by the MLCommons organization that allows datasets to be better described via schema.org \cite{schemaorg}. Croissant hence allows them to be indexed, versioned, and thus findable and reproducible, respectively. The Croissant standard is widely adopted by the machine learning community, and most data repositories and machine learning frameworks support it. In this way discoverability of FeTS++ datasets can be further magnified reaching out to a larger community.
    
 \subsubsection{Data Pre-processing}
    The pre-processing pipeline applied to every case follows the one used across the BraTS and FeTS challenges. Raw scans are converted from DICOM to NIfTI \cite{cox2004sort} and re-oriented to LPS/RAI orientation. All sequences (T1, T1-Gd, T2, T2-FLAIR) are then registered to the SRI-24 atlas \cite{sri} and resampled to 1 $mm^3$. Registration proceeds as follows: N4 bias-field correction is applied temporarily to estimate the transformation matrices and is then discarded, because it otherwise obliterates the tumor signal \cite{bakas2017advancing}. T1-Gd is rigidly registered to the SRI-24 atlas, and then T1, T2, and T2-FLAIR are rigidly registered to T1-Gd. The resulting deformation matrices are then joined and applied to the LPS-oriented scans. After registration, brain extraction removes non-brain tissue such as the skull, neck fat, and eyeballs, using a deep-learning method explicitly developed and evaluated on scans with diffuse glioma and meningioma \cite{thakur2020brain}. All scans are then manually checked to confirm that the whole brain is retained and all non-brain tissue is removed. 
     
\subsubsection{Annotation Protocol}
    All cohorts were annotated following the BraTS annotator-approver model. A pre-defined clinically approved annotation protocol was provided to all clinical annotators, describing the radiologic appearance each tumor sub-region should delineate on the given MRI sequences. Glioma cases were annotated by 60 clinical neuroradiologists volunteering through the American Society of Neuroradiology (ASNR) and the American Society of Functional Neuroradiology (ASFNR). Each case was assigned to an annotator-approver pair, with the two approvers being board-certified neuroradiologists with more than 13 years of experience.
    Annotators could use their tool of preference and follow either a complete manual approach or a hybrid approach in which an automated method produced initial annotations that were then refined manually.
    Each annotation was reviewed by the approver against the corresponding MRI scans and returned to the annotator for refinement until it reached satisfactory quality for sign-off. Meningioma cases followed the same process, with nnU-Net pre-segmentations refined by ASNR/ASFNR volunteers and approved by a board-certified neuroradiologist~\cite{labella2024}. Sub-Saharan African cases were refined by local radiology residents and radiologists before final approval by an expert BraTS neuroradiologist~\cite{adewole2025}. No aggregation of multiple annotations were performed.
    
    The radiologic appearance of each tumor sub-region in MRI, communicated to the radiologists, are as follows:
    \begin{enumerate}
        \item the whole tumor (farthest extent, including the peritumoral edema) delineates the hyperintense abnormal signal on T2 and T2-FLAIR;
        \item the tumor core (TC), including the enhancing, non-enhancing, and necrotic tumor, delineates regions of lower T2 signal and represents the resectable bulk of the tumor;
        \item the enhancing tumor (ET) delineates the hyperintense signal on T1-Gd, after excluding the vessels;
        \item the necrotic core (NCR) outlines regions that appear dark on both T1 and T1-Gd (necrosis/cysts), and regions dark on T1-Gd but bright on T1 (hemorrhage).    
    \end{enumerate}
    Segmentation labels follow the BraTS 2023 convention: label 0 for background, 1 for NCR, 2 for ED, and 3 for ET.
    The tumor entity label (glioma or meningioma) was derived from the source cohort of each case. Glioma cases were histopathologically confirmed diagnosis~\cite{baid2021}, except in the Sub-Saharan Africa cohort, where diagnosis was radiological~\cite{adewole2025}. Meningiomas were diagnosed by imaging or histopathology~\cite{labella2024}.

    The testing datasets
    are not shared with the challenge participants or the public.
    The FeTS collaborators collect and process data from their clinical sites following the pre-processing and annotation protocols presented above. From all these data, approximately 80\% is kept for federated training (outside the context of this challenge) and the rest is set aside for testing and that will be used to evaluate the performance of submitted algorithms. 
    
\subsection{Code Framework}
\label{sec:framework}

    \subsubsection{Software Stack}
    The challenge is delivered as a simulator-first FL environment built on NVIDIA FLARE (NVFlare)\footnote{\url{https://github.com/NVIDIA/NVFlare}}, an open-source, extensible framework developed by NVIDIA~\cite{roth2022nvidia}. NVFlare coordinates the server and participating clients, distributes the global model, and collects local updates for aggregation. Its modular architecture separates federated orchestration from the underlying model and training code, allowing participants to investigate aggregation strategies and local training behavior within a common execution framework. The NVFlare Client API connects the local training code to this workflow through \texttt{flare.receive()} and \texttt{flare.send()}. The framework also supports deployment across independently operated sites, providing a path for subsequent evaluation beyond the challenge’s controlled simulation environment.
    
    Public local evaluation and official hidden evaluation share a single code path and a single frozen evaluator, so that only the data directory differs between them.
    The data pipeline and the baseline model are provided through MONAI\footnote{\url{https://github.com/Project-MONAI/MONAI}}~\cite{cardoso2022monaiopensourceframeworkdeep}, which supplies the image transforms, sliding-window inference, the metric, and a baseline model wrapper.

    \subsubsection{Locked Editing Policy}
    To keep any improvement attributable to the FL method rather than to the model, the baseline model definition is locked.
    Participants edit exactly two files: \texttt{client.py}, which exposes \texttt{local\_train()}, and \texttt{aggregator.py}.
    Every other file is held fixed and checked against a SHA256 manifest at submission time.

    \textbf{Three validity criteria apply:} 1) the model architecture must remain unchanged, 2) each simulated client may access only its assigned institutional data partition, and 3) communication must be restricted to the FLModel payloads permitted by the challenge and exchanged through \texttt{flare.send()} and \texttt{flare.receive()}. These payloads may contain model parameters or parameter differences, metrics, and protocol metadata, but must not carry raw images, reference annotations, or other case-level records. No additional communication channels are permitted.
    
\begin{figure}[t]
        \centering
        \includegraphics[width=\textwidth,alt={Schematic of one federated round in the NVFlare simulator: the server broadcasts the global model, each client trains it locally and returns an update via flare.send(), and the server aggregates the accepted updates into the next global model.}]{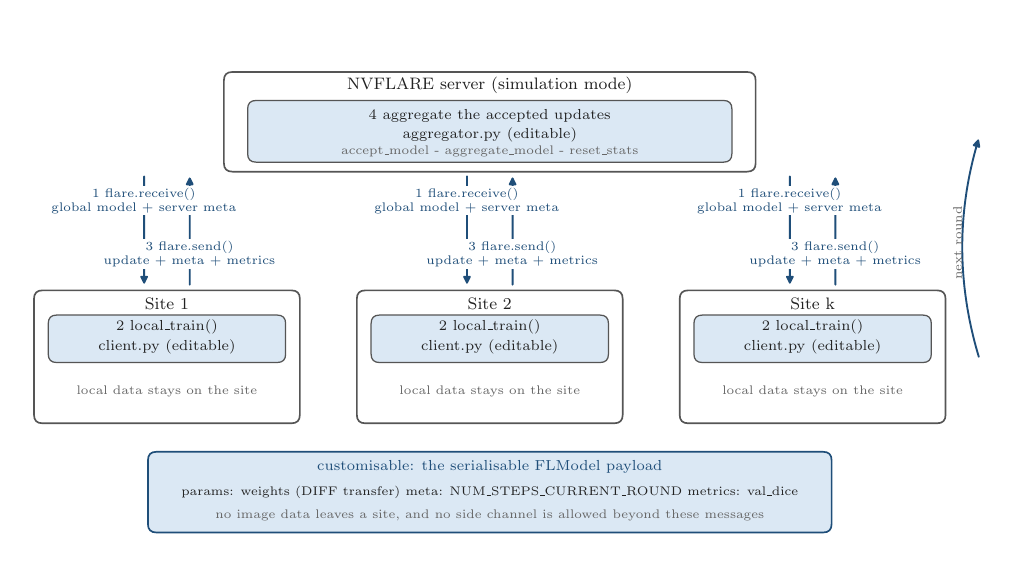}
        \caption{One federated round in the FeTS++ 2027 NVFlare simulator. The server broadcasts the global model, each client receives it with \texttt{flare.receive()}, trains it locally with the participant's \texttt{local\_train()}, and returns the updated model with \texttt{flare.send()}. The server then aggregates the accepted updates into the next global model, and the cycle repeats while \texttt{flare.is\_running()}. Blue elements are participant-editable, namely the local training, the aggregation, and the serializable FLModel payload shown at the bottom, which is the only information that crosses the link. Schematic drawn from the challenge starter code, namely the client runner and the participant aggregator. Each simulated site represents a logical client associated with an institutional data partition. The figure illustrates the challenge’s federated execution protocol within the simulator.}
        \label{fig:nvflare_round}
    \end{figure}

    \Cref{fig:nvflare_round} summarizes one federated round and the two calls that carry it.

\subsubsection{Efficiency}
    Federated methods may trade accuracy against communication cost and overall FL training time, so the challenge scores efficiency jointly with task performance.
    Efficiency is measured as the area under the performance-versus-time (or round) curve, which captures how quickly a method reaches its best accuracy rather than only the final score.
    Efficiency is normalized relative to the best-performing method and carries 20\% of the final ranking, with the full formula given in \Cref{sec:task1_eval} and \Cref{sec:task2_eval}.

\section{Challenge Tasks}
\subsection{Task 1: Federated Tumor Segmentation} \label{sec:task1}
\subsubsection{Task Description}
    Task 1 asks participants to find the best way to aggregate knowledge from segmentation models trained independently at individual institutions, rather than to design a better segmentation network.
    The task is limited to the methodological parts of FL, i.e., 1) aggregation, 2) client selection, 3) per-round local training, 4) compression, and 5) communication efficiency.

    The architecture, data pipeline, splits, initial checkpoint, and federated lifecycle are fixed by the organizers, and the evaluation is hidden while respecting the FL paradigm.
    Any measured difference between submissions therefore reflects the FL method, not model capacity or data access.
    Participants work within the two-file editing surface and validity criteria in \Cref{sec:framework}, and may start from the reference aggregators shipped with the baseline: weighted FedAvg, coordinate-wise median, and clipped mean.

    The FeTS++ 2027 cohort is deliberately realistic rather than balanced: per-site training counts span two orders of magnitude, and the entity mix is strongly site-dependent.
    This makes aggregation non-trivial and connects the task to established results on heterogeneous client data \cite{mcmahan2017communication,kairouz2019advances}.

\subsubsection{Model Definition and Data}
\begin{figure}[htbp]
        \centering
        \includegraphics[width=\textwidth,alt={Bar chart of training cases per site, ordered by decreasing size, with per-site counts spanning 4 to 542 cases and a dashed line marking 50 cases.}]{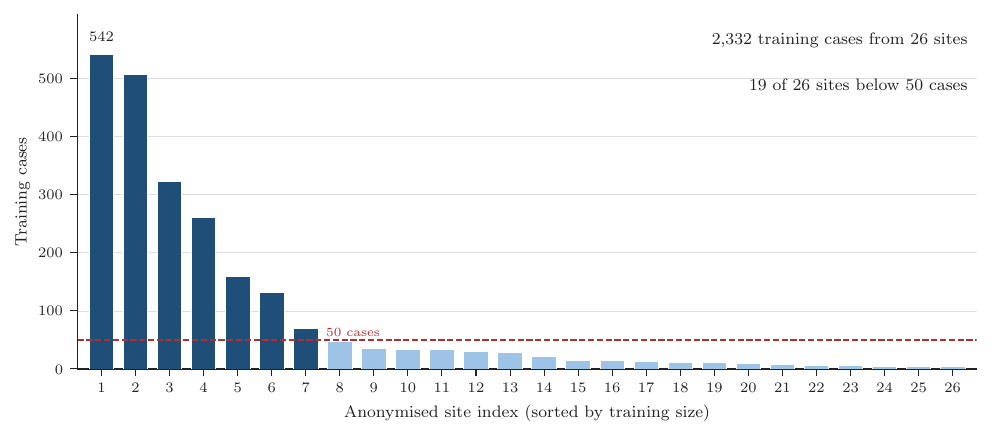}
        \caption{Training cases per site in the FeTS++ 2027 cohort. Bars are ordered by decreasing size and the sites are renumbered for presentation, so the indices do not correspond to the identifiers in the released metadata. The dashed line marks 50 cases, below which 19 of the 26 sites fall, and per-site counts span 4 to 542 cases.}
        \label{fig:task1_cohort}
    \end{figure}

    Task 1 uses a fixed MONAI SegResNet baseline \cite{cardoso2022monaiopensourceframeworkdeep, myronenko20183d} with four input channels, three output channels, an initial filter count of 8, and a dropout probability of 0.2.
    The baseline checkpoint and data pipeline are provided and locked.
    The training set comprises the four co-registered mpMRI sequences and their segmentation masks, together with a partition CSV that assigns each case to its original acquisition institution.
    The cohort is deliberately imbalanced across 26 institutions: per-site case counts range from 4 to 542, and 19 of the 26 sites contribute fewer than 50 cases each.
    \Cref{fig:task1_cohort} shows how unevenly the training cases are distributed across the 26 sites.

\subsubsection{Performance Evaluation} \label{sec:task1_eval}
    Submissions will be scored on the following regions:
    \begin{enumerate}
        \item the ``enhancing tumor'' (ET), label 3;
        \item the ``tumor core'' (TC), labels 1 and 3;
        \item the ``whole tumor'' (WT), labels 1, 2 and 3.
    \end{enumerate}
    The Dice similarity coefficient (DSC) measures the overlap between the predicted mask ($PM$) and the ground truth ($GT$) \cite{dice}:
    \begin{equation}
    \label{eq:dice}
        DSC = \frac{2|GT \cap PM|}{|GT|+|PM|} \, .
    \end{equation}

    The Normalized Surface Distance (NSD) is used to capture boundary agreement between the prediction and the reference: the boundary voxels of each segmentation that lie within a small distance tolerance of the other boundary are counted as matching, so that minor annotation imprecision is not penalized \cite{JiaZhi_MultiConsortium_MICCAISAT2026,Maier_Hein_2024}.
    NSD lies in $[0,1]$, where higher values indicate better boundary agreement, so it complements DSC by focusing on the tumor boundary rather than on the enclosed volume.

\subsection{Task 2: Federated Tumor Classification}
\label{sec:task2}

\subsubsection{Task Description}
    Task 2 asks participants to classify each case as glioma or meningioma under the same FL protocol as Task 1, using the same multi-institutional mpMRI data but without the segmentation masks.
    Correct entity identification is the first clinical decision that conditions the treatment pathway, and it is exactly the kind of label that is unevenly represented across institutions.
    Participants use the same two-file editing surface and validity criteria as Task 1 \Cref{sec:framework}.

    The headline difficulty is that the entity label is almost perfectly confounded with the site of origin, so a training algorithm that simply minimizes an aggregate loss aligns with the majority entity and can degrade on minority-entity clients.
    This is the classical label-skew problem in FL \cite{kairouz2019advances}, here with clinical rather than synthetic provenance.
    Aggregation methods are therefore evaluated not only on average accuracy but on how they behave when the label distribution differs across clients, which makes weighting schemes, client selection, and per-entity calibration all relevant.

\subsubsection{Model Definition and Data}
\begin{figure}[htbp]
        \centering
        \includegraphics[width=\textwidth,alt={Stacked bar chart of the per-site entity composition, showing glioma and meningioma case counts for each of the 26 sites, with 16 glioma-only and 2 meningioma-only sites marked.}]{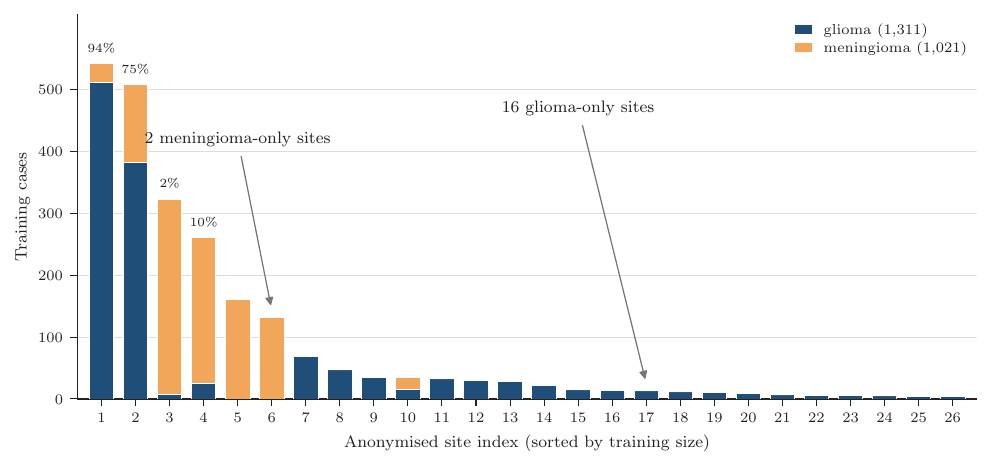}
        \caption{Entity composition per site in the FeTS++ 2027 training cohort. Each bar is one site, split into glioma and meningioma cases, in the same renumbered order as \Cref{fig:task1_cohort}. 16 sites are glioma-only and two are meningioma-only, as marked by the arrows, and the percentages above the 4 largest mixed sites give their glioma share. Sites are renumbered for presentation, and the indices do not correspond to the identifiers in the released metadata.}
        \label{fig:task2_skew}
    \end{figure}

    The classification baseline is fixed and provided. The training inputs are the same four co-registered mpMRI sequences as Task 1, without the segmentation masks, together with a partition CSV and a label CSV that will be released after the segmentation task opens.
    The entity label is heavily confounded with the site: 18 of the 26 training sites are entity-pure (16 glioma-only, 2 meningioma-only), and the mixed sites are heavily unbalanced.
    \Cref{fig:task2_skew} shows the same distribution as the per-site entity composition.

\subsubsection{Performance Evaluation} \label{sec:task2_eval}
    Submissions will be scored with balanced accuracy and the Matthews correlation coefficient (MCC), two metrics that are insensitive to the class imbalance of the cohort.
    Balanced accuracy is the mean of the per-class recalls, and MCC measures the chance-corrected agreement between the predictions and the labels.
    Each metric is ranked across methods, higher being better, with ties broken by rank averaging, and each rank is normalized to a score in $[0,1]$ by linear rescaling of the mean rank across submissions.
    The final score combines balanced accuracy, MCC, and the efficiency term from \Cref{sec:framework} as $S_i = 0.4\,P_{BA} + 0.4\,P_{MCC} + 0.2\,E_i$, where $P_{BA}$ and $P_{MCC}$ are the normalized ranks of the two metrics and $E_i$ is the normalized convergence score of method $i$.
    Accuracy, AUC-ROC, F1, sensitivity, and specificity are reported as diagnostics.

\section{Discussion and Conclusion}

    FeTS++ 2027 provides a comprehensive FL benchmark environment and data that spans across two clinical tasks, tumor segmentation and tumor classification, over an international multi-institutional cohort that is deliberately realistic rather than balanced.
    The shared, locked protocol, under which only the aggregation and client-training logic may change, isolates the contribution of the FL method from the contribution of the model architecture.

    Adding classification alongside segmentation stresses a different property. Because the entity label is confounded with the site, Task 2 tests whether aggregation methods can serve clients whose case mix differs sharply from the global average. The results will show whether the methods that stabilize segmentation also transfer to classification, or whether the two tasks demand different methodological approaches.

    This manuscript describes the challenge design and deliberately reports no results. The results and a complete meta-analysis will follow in a joint publication following the completion of the challenge.

\section*{Acknowledgments}
    
    Research reported in this publication was partly supported by the National Institutes of Health (NIH) under award numbers NIH/NCI:U01CA242871 and NIH/NCI:U24CA279629. The content of this publication is solely the responsibility of the authors and does not represent the official views of the NIH.

\bibliographystyle{ieeetr}
\bibliography{bibliography.bib}

\end{document}